\documentclass[conference]{IEEEtran}

\usepackage{graphicx}
\usepackage{amsmath}
\usepackage{cite}
\usepackage{adjustbox}
\usepackage{float}
\usepackage{makecell}
\usepackage{tabto}
\usepackage{listings}
\usepackage{color}
\usepackage{hyperref}
\hypersetup{colorlinks=false, hidelinks}

\definecolor{green}{rgb}{0,0.7,0}
\definecolor{blue}{rgb}{0,0,0.5}
\definecolor{darkred}{rgb}{0.6,0,0.2}
\definecolor{violet}{rgb}{0.2,0,0.6}

\title{A Comparative Study on Code Generation
with Transformers}

\author{
  \IEEEauthorblockN{Namrata Das\IEEEauthorrefmark{1}, Rakshya Panta\IEEEauthorrefmark{2}, Neelam Karki\IEEEauthorrefmark{3},  Ruchi Manandhar\IEEEauthorrefmark{4}, Dinesh Baniya Kshatri\IEEEauthorrefmark{5}}
  \IEEEauthorblockA{Department of Electronics and Computer Engineering,\\ IOE, Thapathali Campus,\\ Tribhuvan University, Nepal\\
  Email: \{tha075bei026\IEEEauthorrefmark{1}, tha075bei029\IEEEauthorrefmark{2},  tha075bei027\IEEEauthorrefmark{3},
tha075bei035\IEEEauthorrefmark{4}\}@tcioe.edu.np}dinesh.kshatri@thc.tu.edu.np\IEEEauthorrefmark{5}
}

\begin{document}
\maketitle
\begin{abstract}
In an era of widespread influence of Natural Language Processing (NLP), there have been multiple research efforts to supplant traditional manual coding techniques with automated systems capable of generating solutions autonomously. With rapid research for code generation and a sole focus on large language models, there emerges a need to compare and evaluate the performance of transformer architectures based on several complexities of the model. This paper introduces the concept of a ”A Comparative Study on Code Generation with Transformers,” a model based on Transformer architecture, and NLP methodologies to automatically generate C++ source code for different varieties of problems. Here, a comparative study is performed to evaluate the robustness of transformer-based models on the basis of their architecture complexities and their capability to handle diverse problem sets, from basic arithmetic to complex computations.
\end{abstract}

\par
\section{Introduction}
Computer programming has become one of the foundations of technological evolution. With the rapid progress in machine learning and natural language processing, computer programming has become way easier with the help of numerous models capable of generating functional and executable code on the fly. With this research, we attempt to compare the performances and analyze the code execution capability of the generated output of transformer-based code generation models with different architectural complexities. The main objectives of the research are to train a base transformer for generating C++ source code via pseudo-code of problems related to arithmetic, array, string, and sorting operations; to perform transfer learning on a pretrained transformer model for better performance on pseudo-code to C++ source code conversion; to compare the results obtained from the base transformer and pretrained transformer model and analyze the performance. It must be noted that the comparison of performance also centers on the computational resources required by different models with different architectural designs. Additionally, previous works were primarily focused on converting one line of pseudocode into corresponding C++ code statements. The major issue they suffered was the initialization error. This problem was tackled in this project by using complete program pseudocode as input to the model.
\par

\section{Related Works}
There have been numerous endeavors for automated source code generation. Before natural language processing and large language models, most code generation tasks were carried out using UML statecharts \cite{4626805,Niaz2005AnOA,Moreira2010AutomaticCG}. The utilization of machine learning algorithms were limited to compiler heuristics optimization and parallelization\cite{10.1007/3-540-46148-5_5,10.1145/1543135.1542496}. Initial implementations of natural language processing models for code generation can be seen in the work of Barone et al. where they attempted code generation using Python docstrings\cite{barone2017parallel}. Similarly, HTML code generation using wireframes was conducted by Aşiroğlu et al. \cite{inproceedings}. Later on, transformer based models primarily dominated the code generation domain\cite{vaswani2023attention}. Different models were introduced which were customized for tasks such as code completion, code understanding, code generation and also to support multilingual tasks\cite{liu2020multitask, svyatkovskiy2020intellicode,2021}. Meanwhile, comparison studies were also conducted between different neural network architectures for code generation and filling mask tasks\cite{2021}. Alphacode is one of the popular examples of transformer based code generation model for competitive programming \cite{doi:10.1126/science.abq1158}.\\
\\
\  
With the code generation, a number of metrics for code evaluation were also introduced, CodeBLEU being most popular among them\cite{ren2020codebleu}. It is a modified version of popular machine translation evaluation metrics BLEU score\cite{Papineni2002BleuAM}. CodexGLUE, a benchmark for code generation was established by OpenAI to encourage research in multiple code related tasks including clone detection, defect detection, cloze test and code summarization \cite{lu2021codexglue,chen2021codex}. Also, MCoNala, a benchmark for code generation from multiple natural languages was introduced adding more diversity to the feat of automated source code generation \cite{wang2023mconala}. Recently, numerous code generation researches has been carried out introducing extremely versatile and multitasking models like CodeBERT, CodeT5 and GPT-based models which are gaining popularity for implementation and integration in varied set of use cases of code generation, code completion, code correction, code summarization and so on\cite{feng2020codebert, wang2021codet5, liu2023summary}.\\   
\\
\
The SPoC dataset used in this project was introduced and implemented by Kulal et al. using LSTM encoder and decoder \cite{kulal2019spoc}. Their work was further analyzed and attempted by Kaan et al. using transformer architecture\cite{Kaan2021PseudocodeTC}. It must be noted that previous works were primarily focused to convert one line of pseudocode into corresponding C++ code statements. The major issue they suffered was the initialization error. This problem was tackled in this project by using the complete program pseudocodes as input to the model. 

\raggedbottom 

\par
\section{Data Exploration}
\begin{figure}[H]
 \centering
   \includegraphics[width=\linewidth]{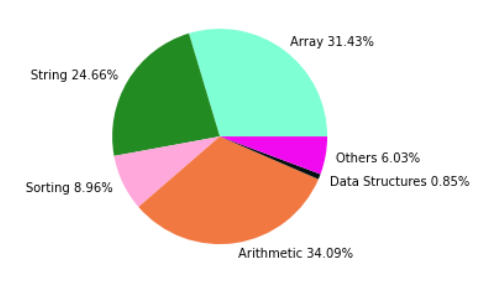}
   \caption{Training Dataset Distribution}
   \label{fig:Training Dataset Distribution}
 \end{figure}

SPoC (Search-based Pseudocode to Code) dataset with 18,356 C++ programs for 677 programming problems having human-authored programs, pseudocodes and test cases, has been used for this project \cite{kulal2019spoc}. Each problem has roughly 27 programs, which are likely to have similar semantics yet different code syntax. The variation in the problem set of the SPoC dataset is shown in Figure \ref{fig:Training Dataset Distribution}.

\begin{figure}[H]
 \centering
   \includegraphics[width=0.85\linewidth, height=0.45\linewidth]{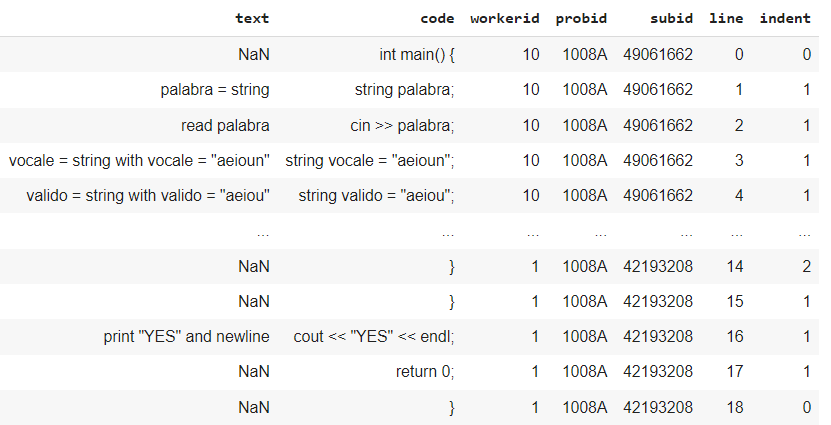}
   \caption{SPoC Dataset Sample}
   \label{fig:SPoC Dataset Sample}
 \end{figure}

\begin{figure}[H]
 \centering
   \includegraphics[width=0.85\linewidth, height=0.35\linewidth]{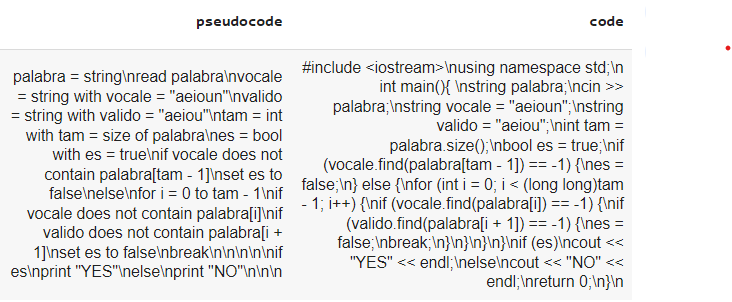}
   \caption{Modified Data Sample}
   \label{fig:Modified Data Sample}
 \end{figure}

In the previous works, the approach taken was to convert one line of pseudocode into corresponding C++ code statement \cite{kulal2019spoc, Kaan2021PseudocodeTC}. This introduced initialization errors. To tackle this issue and to tap into the competency of transformer based models for large sequence of sentences, the dataset has been modified such that the all the code and pseudocode statements belonging to any one program are aggregated together to one input and one reference output to the model. Along with it, code for basic header file imports are also added such that the generated output can be directly used for code generation. The modification brought to the dataset is illustrated in Figure \ref{fig:SPoC Dataset Sample} and Figure \ref{fig:Modified Data Sample}. 

 \setlength{\parskip}{0pt}
 
 \setlength{\parskip}{0pt}

\section{Implementation}
This system implements a structured code generation workflow. The user-entered pseudocode undergoes UTF-8 encoding and is tokenized using a dedicated tokenizer. The parsed and tokenized input text is fed into an encoder, where queries gain similarity by referencing previously stored memory. Employing an attention-weighted mean, the weights are mapped into values, representing stored information. These values are passed through a decoder, detokenized, and then concatenated. The concatenated output tokens form the generated code, which is UTF-8 decoded and postprocessed for optimal UI display. The final result is a C++ output program derived from the user's original pseudocode input.

 \begin{figure}[H]
 \centering
   \includegraphics[width=0.8\linewidth, height=0.7\linewidth]{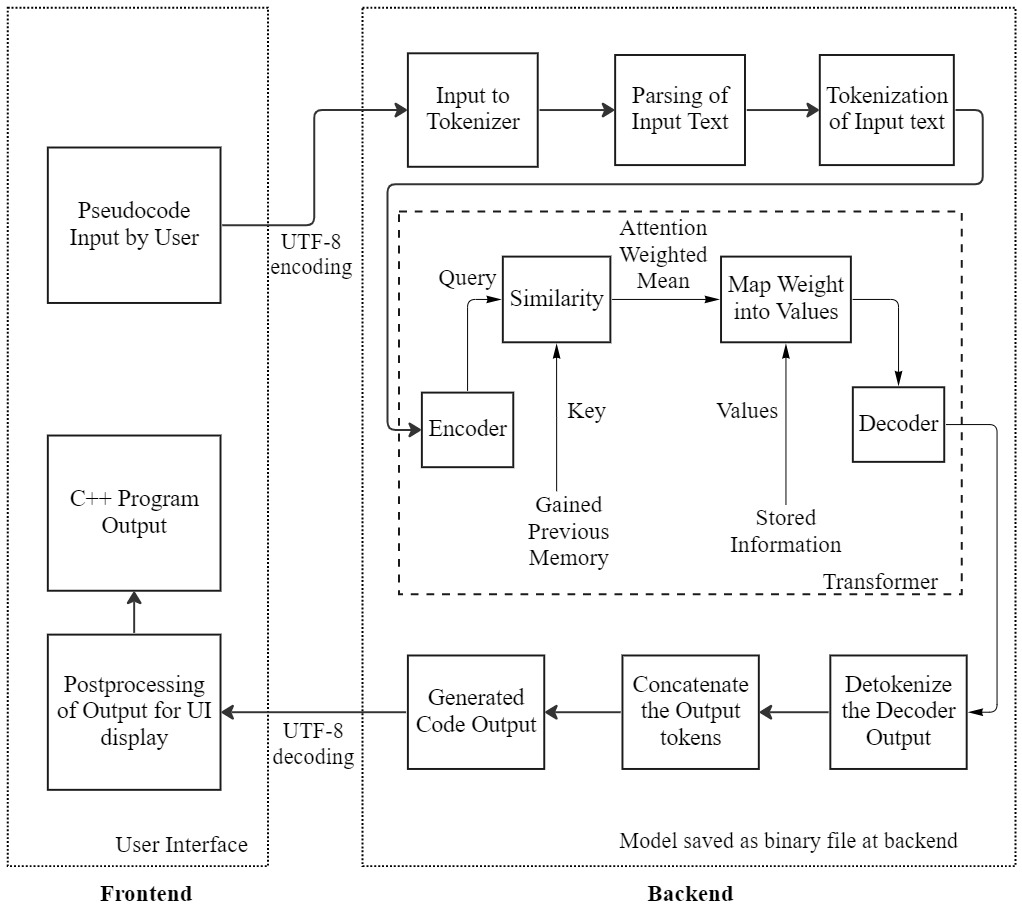}
   \caption{System Block Diagram}
   \label{fig:System Block Diagram}
 \end{figure}
 \setlength{\parskip}{0pt}
\subsection{Transformer Model} 
\begin{figure}[H]
 \centering
   \includegraphics[width=0.8\linewidth, height=0.6\linewidth]{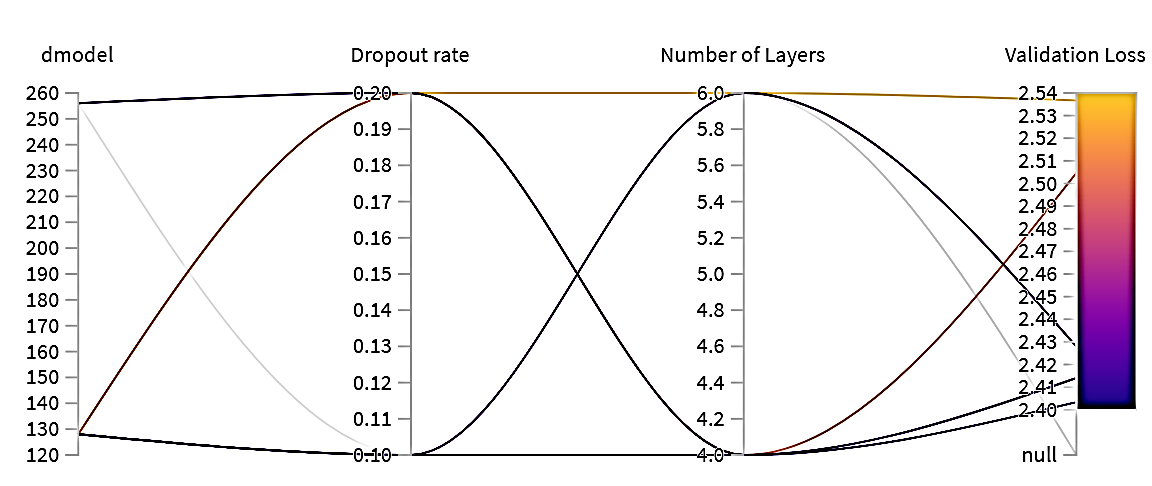}
   \caption{Base Transformer Model Random Search for Hyperparameters}
   \label{fig:Base Transformer Model Random Search for Hyperparameters}
 \end{figure}

A comprehensive process was undertaken to enable the translation of pseudocode to C++ code through a custom-designed Transformer model and a pretrained transformer model fine-tuned on a new dataset and integrated into a web application. For the base transformer model, the BERT tokenizer was fine-tuned and customized to differentiate between reserved and input tokens \cite{devlin2019bert}. This facilitated the conversion of sentences into token IDs, essential for subsequent model input. Upon analyzing the pseudocode and C++ token vocabularies, the study identified 2285 unique tokens for pseudocode and 1989 for C++ programs. Positional encoding vectors with dimensions of 512 were established for 2048 positions.
 \begin{figure}[H]
 \centering
   \includegraphics[width=0.8\linewidth, height=0.6\linewidth]{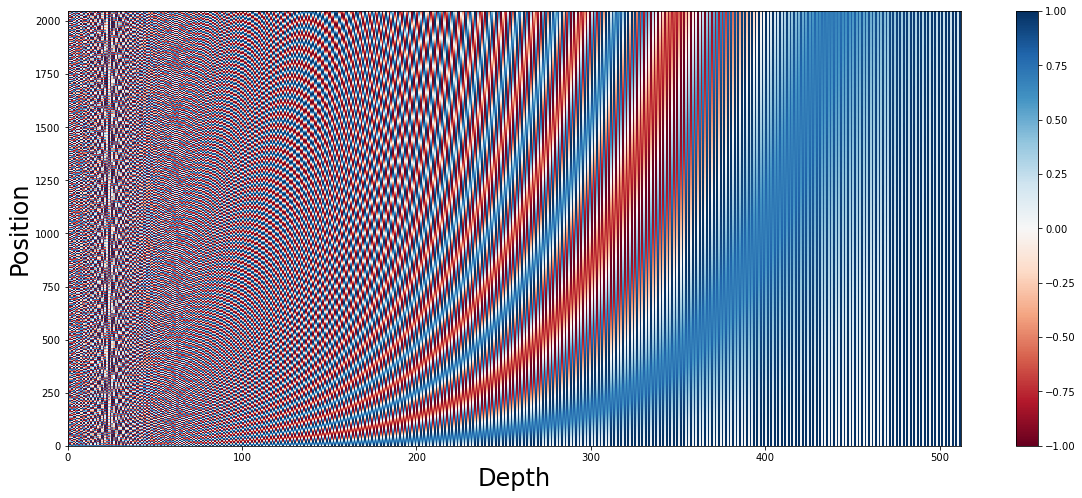}
   \caption{Positional Encoding for Dimension 512 and Sequence Length 2048}
   \label{fig:Positional Encoding for Dimension 512 and Sequence Length 2048}
 \end{figure}
Figure \ref{fig:Positional Encoding for Dimension 512 and Sequence Length 2048} depicts positional encoding vectors with a dimension of 512 for 2048 positions. Blue points indicate positive values, reaching a maximum of +1; white denotes zero; and red points depict negative values, reaching a minimum of -1. 
\\\\
Additionally, masking was employed to disregard padding and look-ahead tokens. Hyperparameters were optimized through 5 iterations of random search, exploring layer counts (4-6), dmodel values (128-256), and dropout rates (0.1-0.2), with selected values of dmodel 128, dropout rate 0.1, and 4 layers for subsequent model training. Upon conducting training for 50, 80, and 30 epochs consecutively, it was noted that the model exhibited signs of overfitting during the 50 and 80 epoch training runs. Consequently, a training duration of 30 epochs with a batch size of 16 was selected as the optimal choice.
\\\\
Figure \ref{fig:Base Transformer Model Random Search for Hyperparameters} shows random search performed to optimize hyperparameters in the base transformer model, including the number of layers, dmodel, and dropout rate. Five iterations were conducted with varying values, and the lowest validation loss occurred with parameters (128, 0.1, 4), resulting in a loss of 2.403 that was selected for model training.

Comparing the base transformer model with pretrained counterparts, the SPoC dataset was fine-tuned on 6 layers of CodeT5-small, consisting of 60.5 million parameters, for pseudocode-to-C++ translation \cite{wang2021codet5}. CodeT5 had been trained for code translation tasks between Java and Csharp. On the grounds of semantic and syntactic similarity between Java and C++, the model was fine-tuned on the SPoC dataset. This further expanded the scope of the evaluation of the code generation capabilities of large language models. Tokenization of text utilized RobertaTokenizer, producing 'input\_ids' and 'attention\_mask' \cite{liu2019roberta}. Dataset preprocessing led to the creation of a data dictionary with batch size of 8 for training and 4 for validation and test sets respectively.
\\\\To optimize performance, a random search explored hyperparameters including learning rate (1e-3 to 5e-5) and warmup steps (500, 1000, 1500). Among five instances, the best results emerged at a learning rate of 8e-4 and 1000 warmup steps, with validation losses ranging from 0.0857 to 0.1116. 

  \begin{figure}[H]
 \centering
   \includegraphics[width=0.8\linewidth, height=0.55\linewidth]{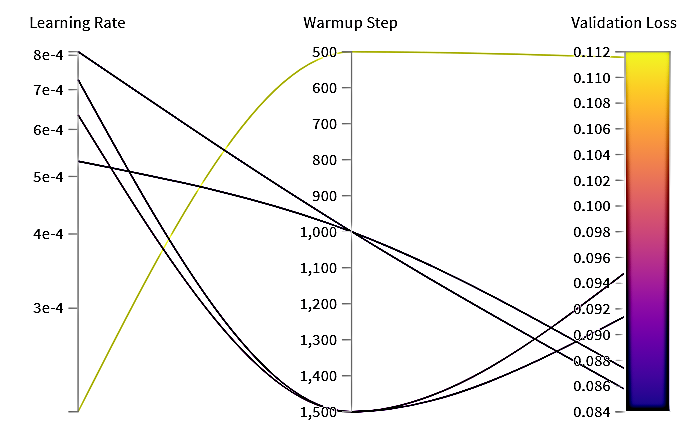}
   \caption{CodeT5 Random Search for Hyperparameters}
   \label{fig:CodeT5 Random Search for Hyperparameters}
 \end{figure}
\begin{figure}[H]
    \centering
    \includegraphics[width=0.7\linewidth, height=0.4\linewidth]{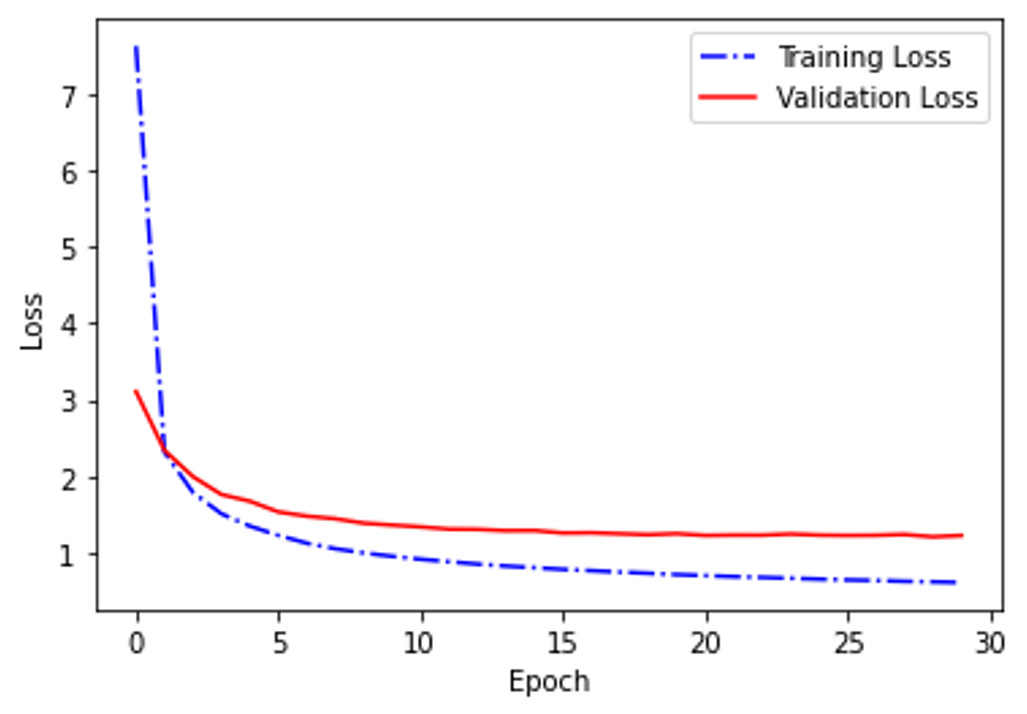}
    \caption{Training and Validation Loss of Base Transformer Model}
    \label{fig:train_val_loss_vanilla}
\end{figure}

\begin{figure}[H]
  \centering
    \includegraphics[width=0.7\linewidth, height=0.4\linewidth]{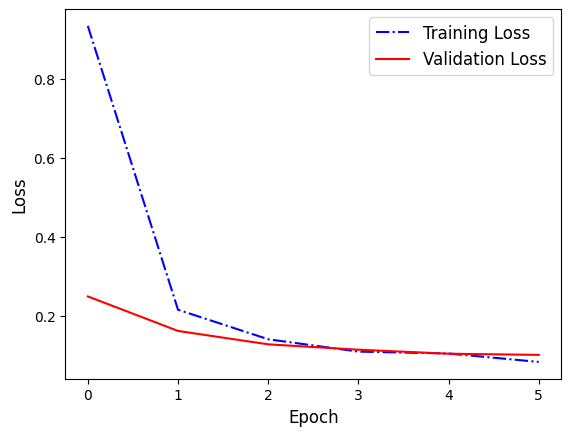}
   \caption{Training and Validation Loss of Fine-tuned CodeT5-small Model}
    \label{fig:train_val_loss_codet5}
 \end{figure}

 
Figure \ref{fig:CodeT5 Random Search for Hyperparameters} shows the random search conducted to optimize hyperparameters in the CodeT5 transformer model, focusing on warmup steps, number of epochs, and learning rate. With the number of epochs fixed at one, five iterations were conducted with varying values of learning rate and warmup steps. The lowest validation loss was observed for the third iteration, with a learning rate of 8e-4 and 1000 warmup steps. Consequently, the model was trained for 5 epochs using a learning rate of 8e-4 and 1000 warmup steps.
\\\\
The implementation leveraged Google Colab's GPU (12.68 GB of RAM) for model training, which took approximately 2 hours. A disk space of 78.19 GB was accessed during GPU. The BERT encoder model facilitated text tokenization. Keras from TensorFlow handled sequential processing, positional encoding, and detokenization in the base transformer model. PyTorch was utilized to load the CodeT5 model into Django and display output in the user interface.
\\\\
To leverage the advantage of reducing memory and computing time by only using a single integer for a class as opposed to a whole vector, these models are trained using the Sparse Categorical Cross Entropy Loss function, which measures the difference between the predicted probability distribution and the actual distribution (ground truth). Cross-entropy is defined as:
\begin{equation}
L_{\text{CE}} = -\sum_{i=1}^{n} t_i \log(p_i)
\end{equation}
where,
\begin{align*}
t_i & : \text{Truth label for the } i\text{th class}, \\
p_i & : \text{Softmax Probability for the } i\text{th class}.
\end{align*}
Figure \ref{fig:train_val_loss_vanilla} shows training and validation loss for the base transformer model trained for 30 epochs. Figure \ref{fig:train_val_loss_codet5} and shows training and validation loss respectively, for the CodeT5-small model trained for 5 epochs.

    \begin{table}[H]
    \centering
        \caption{Model Loss}
        \label{tab:model_loss_table}
        \begin{adjustbox}{width=0.49\textwidth}

        \begin{tabular}{|c|c|c|c|}
            \hline
               \textbf{Models} & \textbf{Metrics} & \textbf{Initial Value} & \textbf{Final Value}  \\  \hline
            Base Transformer & Training Loss & 7.3541 & 0.6095 \\ \cline{2-4}
                        & Validation Loss & 2.7669 & 0.772 \\ \hline
            CodeT5-small Transformer & Training Loss & 5.488 & 0.059 \\ \cline{2-4}
                                    & Validation Loss & 0.2489 & 0.1008 \\ \hline
        \end{tabular}
                \end{adjustbox}

    \end{table}

\par

\section{Results}
In comparison, the results obtained from the CodeT5 small model showed better performance than the base transformer for complex problem sets. However, much difference was not seen for simpler arithmetic programs.

\begin{center}
    \begin{table}[H]
        \begin{small}
            \centering
            \caption{Performance of Models} 
            \label{tab:model_performance_table}
            \begin{adjustbox}{width=0.49\textwidth}
                \begin{tabular}{|c|c|c|}
                    \hline
                     \textbf{Models}  & \textbf{Base Transformer} & \textbf{CodeT5 Transformer} 
                     \\  \hline
                     Similarity Score & 0.1461 & 0.8799
                     \\ \hline
                     CodeBLEU & 0.1892 & 0.8659 
                     \\ \hline
                     Ngram Match Score & 0.272 & 0.865 
                     \\ \hline
                     Weighted Ngram Match Score & 0.162 & 0.849 
                     \\ \hline
                     Syntax Match Score & 0.2375 & 0.8519
                     \\ \hline
                     Dataflow Match Score & 0.4758 & 0.8981
                     \\ \hline
                \end{tabular}
            \end{adjustbox}

        \end{small}
    \end{table}
\end{center}

Models’ performance is evaluated using various metrics that capture different aspects of code generation quality. Let’s delve into the explanations of these metrics: \\

i. BLEU Score: BLEU (Bilingual Evaluation Understudy), a metric commonly used in NLP (Natural Language Processing) tasks, calculates the precision of n-grams in the generated output concerning the reference output. A higher BLEU score indicates better similarity.
\begin{multline}
\text{BLEU} = \min\left(1, \exp\left(1 - \frac{\text{reference length}}{\text{output length}}\right)\right) \\
\times \left(\prod_{i=1}^{4} \text{precision}_i\right)^{\frac{1}{4}}
\end{multline}

where,
\begin{align*}
precision_i & : \text{precision at } i\text{-gram}, \\
m & : \text{No. of candidate translation words} \\ & \text{occurring in reference translation}, \\
w_t & : \text{Total no. of words in the candidate translation}.
\end{align*}

ii. CodeBLEU: Extending the BLEU score to code generation tasks, considering not only token precision but also the structural and syntactic correctness of the generated code, CodeBLEU provides a more tailored evaluation for code-related tasks. \\

iii. N-gram Match: This metric measures the proportion of matched n-grams between the generated code and the reference.
\begin{equation}
\text{N-gram Match} = \frac{\text{Total Number of n-grams in Reference}}{\text{Number of Matched n-grams}}
\end{equation}\\

iv. Weighted N-gram Match: Considering not only the proportion of matched n-grams but also the significance of each match, this metric assigns weights to each matched n-gram based on their importance. \\
\begin{multline}
\text{Weighted N-gram Match} = \\
\sum_{n=1}^{N} w_n \times \frac{\text{Number of Matched } n\text{-grams}}{\text{Total Number of } n\text{-grams in Reference}}
\end{multline}

where,
\begin{align*}
w_n & : \text{Weight of } n\text{-grams}
\end{align*}

v. Syntax Match Score: It evaluates the syntactic correctness of the generated code in comparison to the reference. The syntactic match score assesses how well the generated code adheres to the syntactic structures of the target programming language, C++, in this case.\\

vi. Dataflow Match: It assesses the alignment of data flow structures between the generated and reference code, accounting for the information passed and manipulated within the code. A higher dataflow match score indicates a closer correspondence in the data flow structures between the generated and reference code. \\

In Table III, additional white spaces can be observed in the output of the base transformer model, owing to punctuation marks treated as separate tokens by the model. The white spaces hinder the generated output from being used directly for execution. The 'return 0' statement was not found in the output of CodeT5. \\

\begin{table}[H]
   \centering
   \label{tab:model_output}
    \caption{Output of Base Transformer and CodeT5 Transformer}
    \begin{adjustbox}{width=0.49\textwidth}
    \tiny
  \begin{tabular}{|c|c|c|c|}
    \hline
    \textbf{\thead{Type of\\Problem}} & \textbf{\thead{Pseudocode}} & \textbf{\thead{Output by\\Base Transformer}} & \textbf{\thead{Output by\\CodeT5 Transformer}} \\

    \hline
    Addition 
    & \makecell[tc]{ declare a,b \\ input a,b \\ add a,b}
    & \makecell[tc]{
     \#include $<$iostream$>$ \\
    using namespace std ; \\
    int main ( )\{ \\
    int a , b ; \\
    cin $>>$ a $>>$ b ;\\
    cout $<<$ a + b $<<$ endl ;\\
    return 0 ;\\
    \}
    }
    & \makecell[tc]{ 
     \#include $<$iostream$>$\\
    using namespace std ; \\
    int main ()\{ \\
    int a, b ; \\
    cin $>>$ a $>>$ b ;\\
    cout $<<$ a + b $<<$ endl ;\\
    \} 
    }\\
\hline

    \makecell[tc]{\\Largest \\ of three \\ numbers}
    & \makecell[tc]{\\ define integers n1, n2, n3 \\ display "enter three numbers" \\ read n1, n2, n3 \\ if n1$>$=n2 and n1$>$=n3 \\ print "largest is " n1 \\ else if n2$>$=n1 and n2$>$=n3 \\ print "largest is " n2 \\ else \\ print "largest is " n3}
    & \makecell[tc]{\\
    \#include $<$iostream$>$ \\
    using namespace std; \\
    int main ( )\{ \\
    int n1, n2, n3; \\
    cin $>>$ n1 $>>$ n2 $>>$ n3 ;\\
    int n3 = n1; \\
    if( n1 $>$= n2) \\
    \{ \\
    cout $<<$ "first" $<<$ endl;\\ 
    \}
    else if( n1 $>$= n2) \\ \{ \\
    cout $<<$ "second" $<<$ endl;\\ \} return 0; \\ 
    \}
    }
    & \makecell[tc]{\\
    \#include $<$iostream$>$ \\
    using namespace std; \\
    int main ()\{ \\
    int n1, n2, n3; \\
    cout $<<$ "enter three numbers" $<<$ endl;\\ 
    cin $>>$ n1 $>>$ n2 $>>$ n3 ;\\
    if( n1 $>$= n2 \&\& n1 $>$= n3) \\
    cout $<<$ "largest is" $<<$ n1 $<<$ endl;\\ 
    else if( n2 $>$= n1 \&\& n2 $>$= n3) \\
    cout $<<$ "largest is" $<<$ n2 $<<$ endl;\\ 
    else \\
    cout $<<$ "largest is" $<<$ n3 $<<$ endl;\\ 
    \}
    } \\
    \hline

    \makecell[tc]{\\Addition \\ of numbers \\ in array}
    & \makecell[tc]{\\ define i,sum \\ define n=5 \\ declare arr[]=\{1,2,3,4,5\} \\ sum=0 \\ for i=0 to i$<$n \\ sum = sum + arr[i] \\ i=i+1 \\ end for \\ print sum}
    & \makecell[tc]{\\
    \#include $<$iostream$>$ \\
    using namespace std ; \\
    int main ( )\{ \\
    int i, n, sum = 0; \\
    cin $>>$ n ;\\
    for (int i = 1;i$<$ =5;i ++) \\ \{ \\
    cin $>>$ arr [i] ;\\
    sum + =  arr[i]; \\ \} \\
    for (int i = 1;i$<$ =5;i ++) \\ \{ \\
    sum + =  arr[i]; \\ \} \\
    cout $<<$ sum $<<$ endl ;\\
    return 0;\\ \}
    }
    & \makecell[tc]{\\
    \#include $<$iostream$>$ \\
    using namespace std ; \\
    int main ()\{ \\
    int i, sum; \\
    int n=5; \\
    int arr[] = \{1, 2, 3,4,5\};
    int sum = 0; \\
    for (int i = 1 i$<$ n i ++) \{ \\
    sum + =  arr[i]; \\ \} \\
    i = i + 1; \\ \} \\
    for (: i$<$ n;i ++) \\ 
    cout $<<$ sum $<<$ endl ;\\ \}
    }\\
    \hline

    \makecell[tc]{\\Generate \\ fibonacci\\ series}
    & \makecell[tc]{\\Declare variable a, b, c, n, i, \\ a=0, b=1, 1=2 \\ Read n from user \\ Print a and b \\ Repeat for i = 2 to 1$<$=n : \\ c = a + b \\ print c \\ a=b \\ b=c \\ end loop \\ end}
    & \makecell[tc]{ \\
     \#include $<$iostream$>$ \\
    using namespace std ; \\
    int main ( )\{ \\
    int n, a = 0, b = 0 ; \\
    cin $>>$ n ;\\
    for (int i = 1;i$<$ = n;i ++) \{ \\
    cin $>>$ a ;\\ b = a; \\ c= b; \\ \} for (int i = 2 ;i $<=$ n;i++) \\ \{ \\
    if (a== b) \\ \{ \\ c=a; \\ \}\} cout $<<$ c $<<$ endl ;\\
    return 0 ;\\
    \}
    }
    & \makecell[tc]{\\
    \#include $<$iostream$>$ \\
    using namespace std ; \\
    int main ()\{ \\
    int a,b,c,n,i; \\
    a=0, b=1, c,n,i;\\
    cin $>>$ n ;\\
    cout $<<$ a $<<$ " " $<<$ b $<<$  endl ;\\
    for (int i = 2;i$<$ = n;i ++) \{ \\
    c=a+b;\\
    cout $<<$ c $<<$ endl ;\\
    a=b;\\
    b=c;\\ \} \\
    break;\\ \}
    }\\
    \hline
  \end{tabular}
  \end{adjustbox}
\end{table}

\par

\section{Discussion}
From the study, it was found that, though the number of layers and training parameters heavily affect the code generation performance, acceptable results can be obtained by transformers with very few layers and training parameters. The model was able to overcome the initialization errors found in previous works of code generation with the SPoC dataset when a complete program pseudocode and its corresponding C++ source code were input as pairs in the model instead of statement-wise input  \cite{kulal2019spoc,Kaan2021PseudocodeTC}. Also, the models trained on different programming languages can generate executable source codes for an unseen programming language due to syntactic similarities. Here, in the case of CodeT5, the model was originally trained for translation between Csharp and Java \cite{wang2021codet5}. The model was fine-tuned on a dataset of pseudocodes and C++ code pairs, and the results obtained were directly executable in most cases.
\\\\
The code generation ability of the base transformer model for complicated problems with multi-step logic execution was not found to be executable and had multiple syntactic errors. The reason can be considered to be the small size and low variation of code problems in the training dataset. The base transformer model's performance seemed to decline with an increase in the complexity of problems, whereas CodeT5's performance was found to be robust enough to handle a variety of problems. The reason can be attributed to the multi-tasking capability of the model with training on a huge dataset of CodeSearchNet \cite{DBLP:journals/corr/abs-2009-07839} with multiple programming languages. It is worth mentioning that the performance of models built on transformer architecture when trained on a smaller dataset requires training for multiple epochs to obtain reasonable output. This results in overfitting of the model on the training set and degrades its ability to handle unseen data as inputs. Hence, dataset sizes can be seen as the bottleneck for transformer models. The model's performance can be improved if the dataset is increased in size and variations in the complexity of problems are introduced. \\\\
It was found that the robustness of multiple code generation models to generate code with few sentence commands is a result of their huge architectural design with multiple layers and huge training data corpus. Along with design, the resources required to train such a large language model, result in significant computational costs. The computational requirements are not limited to training and validation stages, but the operational costs of models during their inference stages are also high. The average time taken to generate code for any input varied from 5 to 15 seconds for CodeT5 and 15 to 60 seconds for the base transformer model, respectively. Time requirements were found to heavily depend on the device used for computation, with faster results on GPUs. For faster results, GPUs with higher computing capabilities are required, which again leads to increased costs. Hence, the integration of large language models into the deployed systems is an expensive approach to improve the systems. 
\\

\section{Conclusion}
It can be concluded that the corpus on which language models are trained heavily affects their performance in generative tasks. Also, with increasing number of layers and parameters, the language models become more robust to handle unseen problems. It was observed that the CodeT5 model, though trained for translation between Csharp and Java, handled code generation tasks for the C++ language quite well when fine-tuned on pseudocode to code translation. The performance was found to be better than the four-layered base transformer model specifically trained for code generation from pseudocodes. The robustness of CodeT5 can be attributed to its large architecture with a higher number of layers than the base transformer and the large corpus of data that it has been trained on to carry out a number of coding tasks. In contrast, the resources and time required to train a few layered base transformer model was quite low in comparison to the resources required for training a large language model with billions of parameters. Hence, cost and performance trade-offs can significantly affect training a large language model for specific tasks.

\section{Acknowledgements}
We would like to extend our deepest gratitude to our project supervisor Er. Dinesh Baniya Kshatri for providing us much needed guidance during this project. We are grateful to the Department of Electronics and Computer Engineering, Thapathali Campus for their help and suggestions in the selection and hopeful execution of this project. Our deepest gratitude to the widely growing machine learning community that provides so many resources for study and research available to learners\cite{Jin2022,Django,Kostadinov2019,NLTKProject,Phi2020,Singh2020,Yeaung2020}.
At last we would like to thank everyone who were directly or indirectly involved for the successful execution of this research.
\\


\end{document}